%% file: main.tex
\documentclass[11pt]{article}

\usepackage[margin=1in]{geometry}
\usepackage{graphicx}
\usepackage{amsmath,amssymb}
\usepackage{bm}
\usepackage{booktabs}
\usepackage{tabularx}
\usepackage{array}
\usepackage{ragged2e}
\usepackage{float}
\usepackage{microtype}
\usepackage[hidelinks]{hyperref}

\newcommand{\TBL}[2]{#1\par\noindent\makebox[\linewidth][c]{#2}\par}
\newenvironment{fntable}[1][\linewidth]{\begingroup\small}{\par\endgroup}

\title{Spatiotemporal Agility: Time-Constrained Reinforcement Learning for Vision-Guided Dynamic Quadrupedal Interception}
\author{
Yidong Zhu\textsuperscript{1},
Zibo Dai\textsuperscript{1,\ensuremath{\dagger}},
Tongning Zhang\textsuperscript{1,\ensuremath{\dagger}},
Leixin Chang\textsuperscript{1}, and
Hua Chen\textsuperscript{2,*}\\[0.6em]
\small \textsuperscript{1}Zhejiang University, Hangzhou, China\\
\small \textsuperscript{2}LimX Dynamics Technology Co., Ltd., Shenzhen, China\\
\small \textsuperscript{\ensuremath{\dagger}}Indicates equal contribution\\
\small \textsuperscript{*}Corresponding author: \href{mailto:huachen.work@gmail.com}{huachen.work@gmail.com}
}
\date{}

\begin{document}

\maketitle

\begin{abstract}
\input{source/abstract}
\end{abstract}

\input{source/introduction}
\input{source/related_work}
\input{source/method}

\input{source/experiment}
\input{source/conclusion}

\section*{Author Contributions}
Yidong Zhu led the reinforcement-learning component, including policy training, reward and parameter tuning, simulation experiments, and real-world robot experiments. Zibo Dai and Tongning Zhang developed and supported the vision/perception component and assisted with real-world experiments, including environment setup and data recording. Leixin Chang provided research guidance, technical supervision, and discussions on the experiments and methodology. Hua Chen supervised the project, provided overall research direction, and guided manuscript preparation.

\section*{Financial Support}
This work was supported by the Physical Intelligence Lab (Phi Lab) at the Zhejiang University--University of Illinois Urbana-Champaign Institute (ZJU-UIUC Institute), including laboratory resources and experimental support.

\section*{Conflicts of Interest}
The authors declare no conflicts of interest exist.

\section*{Ethical Approval}
Not applicable.

\section*{Acknowledgements}
The authors used AI-assisted tools for LaTeX formatting and language-editing support during manuscript preparation. All authors reviewed the manuscript and take full responsibility for its content.

\bibliographystyle{unsrt}
\bibliography{references}

\end{document}

%% file: source/abstract.tex
Legged robots require robust agility to perceive and interact with complex and dynamic environments within a constrained time. However, most existing quadruped locomotion works rely on velocity-tracking policy, which struggle to reach precise targets within strict temporal constraints. Moreover, integrating real-time perception with agile locomotion for highly dynamic targets remains challenging due to sensor latency and processing delays. To concretely study and benchmark such agility in dynamic settings, we introduce a challenging ball-catching task for legged robots. This paper proposes an integrated framework that combines a vision module for landing point and time prediction with  a direct position and time conditioned RL locomotion policy, instead of intermediate velocity commands. Beyond the method design, this work presents a system-level contribution that completes real-time robotic interception system that integrates multi-camera perception, online trajectory prediction, low-latency target communication, and sim-to-real locomotion control into a closed-loop deployment pipeline. By explicitly predicting the future spatial-temporal target, our approach mitigates perception latency during dynamic interception. We conducted extensive ball-catching experiments for the legged robot. Through comparative experiments against a velocity-tracking baseline, our direct target-conditioned approach achieves a higher success rate in catching balls with predicted landing spots within 2 meters and flight times between 0.8 and 1.2 seconds. This shows that the robot has successfully completed the dynamic ball-catching task under our tested setup. Furthermore, our policy exhibits a smaller performance gap after deployment, suggesting improved sim-to-real behavior in these trials.

%% file: source/introduction.tex
 \section{Introduction}
Agility is the ability to rapidly perceive, plan, and execute coordinated whole-body motions under tight time constraints. This is reflected in the robot's extremely reactive performance to specific tasks, such as catching a ball within a short time.
These tasks are difficult for quadruped robots, but easy for animals. For example, dogs have high-fidelity perception, fast prediction of ballistic trajectories, and coordinated whole-body dynamics to intercept thrown objects with apparent ease. For legged robots, reproducing this behavior is not merely a question of locomotion speed: it requires proper whole-body coordination and pushing the limits of sensing latency, state estimation, prediction accuracy, and dynamic control simultaneously. Small errors in perception or timing can render an otherwise agile gait useless for interception, while overly aggressive control can induce instability, falls, or hardware stress. \cite{10161392, roychoudhury2023perception}, use perception-based methods to enhance the real-time performance and motion accuracy. \cite{10801909, 9981198, 10610200, he2024agilesafelearningcollisionfree, Tan-RSS-18, doi:10.1177/02783649241285161} use reinforcement learning (RL) methods to produce agile locomotion performance and obstacle avoidance strategies in challenging terrains. 
\cite{11127596, li2024learning, fu2025learningdiversenaturalbehaviors} combine multi-stage pipelines through multi-skill learning, enhancing task coverage and generalization capabilities. 
However, few studies explicitly tackle the vision-guided, time-aware interception problem.
The current methods usually optimize either velocity tracking, orientation, or energy consumption separately or their linear combination, making it difficult to balance these objectives under strict time constraints. 
The time-constrained positionally goal-conditioned training fashion provides a more holistic and more general target-conditioned approach, opening up a larger space of possible solutions that may lead to the discovery of efficient motion patterns \cite{9981198}. Building such a system remains challenging because perception, prediction, communication, and control operate at different rates and introduce cumulative latency and error. A reliable spatiotemporal interface is therefore essential for interception within an approximately one-second horizon.

In this work, we focus on the position-conditioned, time-aware regime and take the vision-guided dynamic ball interception task as a case of time-constrained dynamic target reaching. We identify and address two practical challenges that are often underestimated in prior RL locomotion work: (1) \emph{temporal misalignment} where policies that minimize spatial error do not necessarily arrive at the correct time for interception and (2) degenerate locomotion which means naive emphasis on rapid arrival can produce pathological gaits, such as base sinking or excessive pitching, or poor omnidirectional responsiveness. To overcome these issues, we integrate a locally coordinated perception pipeline that provides both predicted impact positions and explicit time-to-impact estimates, and we design an RL objective that prioritizes spatial accuracy and arrival-time synchronization while preserving stability and energy efficiency. This specific integration produces a rotation-prioritized interception behavior in our trials, within the tested interception workspace.

The main contributions are summarized as follows:
\begin{enumerate}
    \item We develop a decoupled real-time quadrupedal interception framework that integrates multi-camera perception, landing point and time prediction, target communication, and RL-based position-tracking locomotion into a closed-loop hardware pipeline, addressing the tight coupling between perception and locomotion in time-constrained interception through a modular system design.
    \item We introduce a co-located prediction and locomotion to achieve simultaneous position reaching. By explicitly aligning spatial and temporal objectives, our time-aware, position-conditioned policy operates seamlessly with the predictive vision module
    \item We conducted extensive comparative simulation and real experiments validate the effectiveness of our integrated approach and showcase higher success rates under the tested setup over the traditional velocity-tracking baseline and instantaneous ball-state baseline in dynamic interception scenarios.
\end{enumerate}

%% file: source/related_work.tex
\section{Related Work}
\subsection{Legged locomotion for dynamic mobility}
Legged RL has recently pushed agility, adaptability, and skill composition to new heights. Works like \cite{10801909, 9981198, 10610200} demonstrate that policies can navigate risky and irregular terrains, perform obstacle-traversal, and achieve strong locomotion robustness. Safety and contact constraints have also been incorporated in high-speed locomotion in \cite{he2024agilesafelearningcollisionfree}, while curriculum and multi-skill learning methods, like \cite{li2024learning, fu2025learningdiversenaturalbehaviors, 11127596}, enable flexible switching among behaviors and more generalization. Additionally, recent work on jumping behaviors via RL with impedance matching shows that legged robots can perform large jumps in real hardware, such as 55 cm distance or 38 cm height, while still maintaining stable walking across multiple directions. \cite{10597522}

On the catching or interception side, a number of works combine perception, prediction, and locomotion. For example, \cite{10161392} uses an event camera for very low latency catching; \cite{10801922} enables a quadruped robot’s front legs to catch thrown objects using vision and trajectory prediction; \cite{11127596} trains a mobile manipulator with a dexterous hand to track and then catch in flight. Many of these focus on minimizing perception delay or improving spatial observability, but seldom incorporate explicit time-limited conditions into the policy or shape rewards to enforce synchronized interception. Our work studies a complementary off-board perception and basket-interception setting, rather than claiming to outperform these onboard catching systems.

\subsection{RL for Position-Conditioned Legged Locomotion}
Position-conditioned policies, where the low-level controller receives target poses or positions as commands, have become a standard design choice for modular quadruped control, as they simplify high-level planning while reusing learned primitives across tasks. End-to-end and hierarchical RL approaches have demonstrated strong agility and generalization. For example, \cite{10801909} presented a generalist locomotion policy fine-tuned for risky terrains such as stepping stones and narrow beams, achieving real-world traversal at speeds above 2.5 m/s. Similarly, \cite{9981198} introduced an end-to-end local navigation framework combining locomotion and navigation via dense pose/velocity shaping rewards. Hierarchical skill-selection methods, such as extending locomotion policies with jump \cite{doi:10.1126/scirobotics.adi7566}, crouch, and climb skills to achieve parkour-like agility. Other studies leverage priors, residual learning, or skill adaptation to broaden flat-terrain policies to complex terrain \cite{pmlr-v305-zhang25j, 10610200}, while whole-body loco-manipulation combines locomotion with manipulation by conditioning on task-space poses \cite{10806624}.

Despite these advances, two issues remain underexplored: (1) \textit{temporal misalignment}, where position-conditioned policies lack explicit objectives to synchronize arrival with event timing, and (2) \textit{degenerate locomotion}, where naive position tracking can cause undesirable behaviors such as base sinking or limited omnidirectional responsiveness. Our work explicitly addresses these limitations by augmenting the observation space with time-to-impact information and by designing fused rewards that preserve locomotion stability while ensuring interception timing.

\subsection{Object Detection and Tracking}
Object detection is a foundational step for tracking and subsequent trajectory prediction. A common approach is to mark the target object with a specific color and segment it within the frame by selecting a predefined range in the HSV (hue-saturation-value) color space. Alternatively, neural networks such as YOLOv13n \cite{lei2025yolov13realtimeobjectdetection} can be used to detect and localize the target via bounding boxes. Once 2D image coordinates are obtained, an Intel RealSense RGB-D camera can provide depth information, enabling the computation of the object’s 3D position relative to the camera using intrinsic and extrinsic calibration parameters. In this work, we use YOLOv13n as a recent object detector, along with an RGB-D camera, to achieve robust detection and localization.

Another widely used method involves motion capture systems such as VICON \cite{9341134}, which offer highly accurate 3D tracking through marker-based detection. However, such systems require extensive multi-camera setups and physical markers attached to objects, limiting their flexibility in unstructured environments.

Frame-based cameras are frequently used to follow flying objects in systems such as the Hamlet badminton robot \cite{10801922} and the \textit{“Catch It!”} mobile manipulator \cite{11127596}. These systems typically rely on onboard cameras, which are susceptible to motion blur and occlusion during rapid robot movements. To mitigate these issues, some works propose using event cameras \cite{10161392}, which offer high temporal resolution and reduced latency, enabling tracking at speeds up to 15 m/s. However, many vision-based methods impose specific operational constraints. In contrast, our work adopts a globally coordinated perception system that uses two global cameras for both ball tracking and robot localization. This design decouples perception from robot motion, minimizing issues such as blur and occlusion, and providing reliable state estimates for robot tracking tasks.

\subsection{Dynamic Trajectory Prediction}
Accurately predicting the trajectory of dynamic objects remains a core challenge in robotic interception. Traditional model-based techniques, such as Kalman filters and extended Kalman filters (EKF), form a common foundation by integrating simplified physical dynamics, including gravity, for trajectory propagation. Regression-based methods, notably including \cite{10801922}, are also widely employed. These techniques are computationally efficient and perform reliably under predictable motion models.

With advances in machine learning, data-driven methods have gained prominence. For instance, recurrent neural networks (RNNs) such as the Neural Acceleration Estimator (NAE) \cite{9635983} have been used to learn complex dynamics and improve prediction accuracy. Nevertheless, such approaches typically require large datasets and extensive training.

Some methods seek to increase reaction time by observing the thrower’s motion before object release \cite{10161392}, while others leverage event-based sensing to overcome conventional latency-bandwidth trade-offs. Despite these innovations, many prediction frameworks remain vulnerable to perceptual delays and environmental disturbances. In this work, we use a Kalman filter for trajectory prediction due to its simplicity and adequacy for our task.

%% file: source/method.tex
\section{Method}

Our goal is to develop a framework for quadruped robots that decouples perception and prediction with RL locomotion training. This framework is capable of capturing a moving ball at high speed and calculating the landing position and timing before the ball hits the ground, allowing the robot to catch it successfully. In this section, we describe the vision-based localization and trajectory prediction pipeline, the time-aware RL formulation, and the teacher-student distillation framework used for agile local navigation.

\begin{figure*}[!htbp]
\centering
    \includegraphics[width=0.8\textwidth]{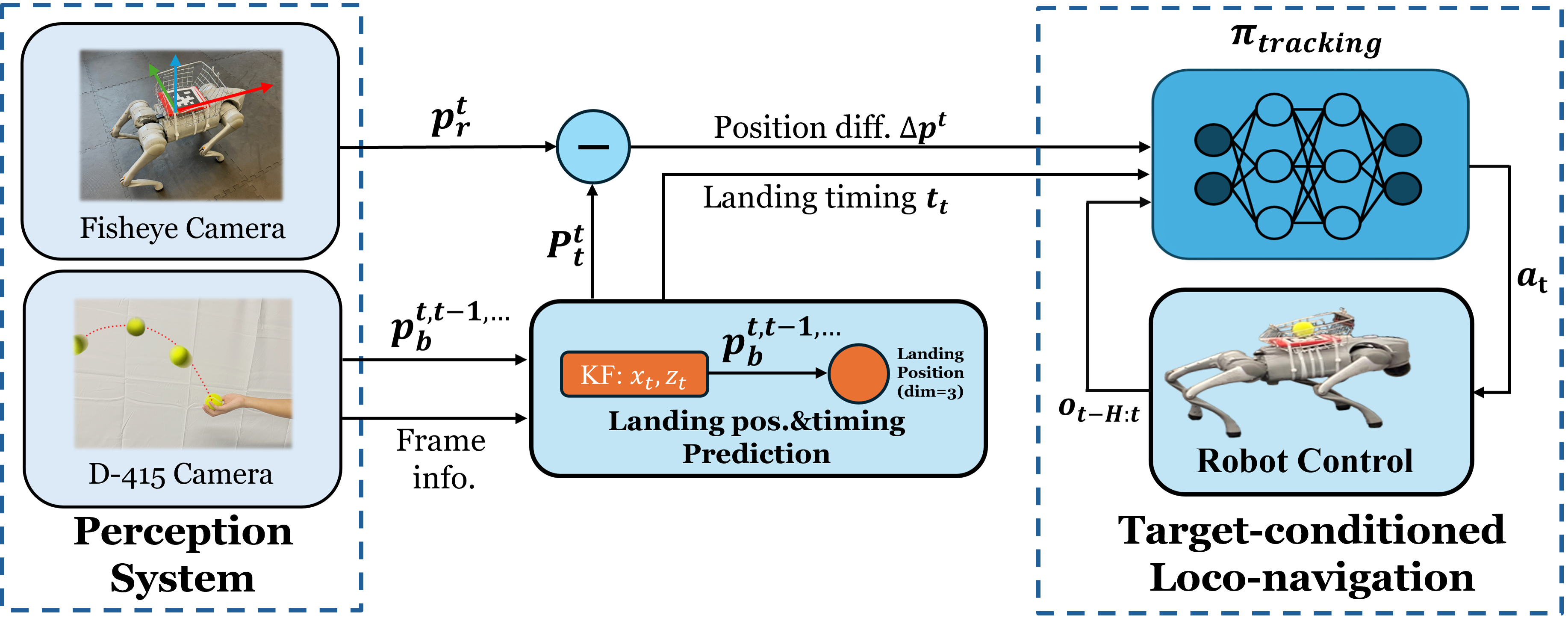}
    \caption{\textbf{System Overview}. Our system comprises a visual part and an RL part. The visual part gets the RGB input of the ball and the April-tag on the back of the robot, and outputs the position of the robot and the estimated ball landing position and time. Then the RL part gets that information as policy input to send the action to the robot}
    \label{fig:pipeline}
\end{figure*}

\input{source/visual}
\input{source/rl_part}

%% file: source/visual.tex
\subsection{Visual Perception and Prediction}

Agility in catching tasks is not only limited by the controller but is strongly constrained by the perception latency and robustness. To ensure that our reinforcement learning (RL) policy receives precise target states under rapid motions, we design a globally coordinated, multi-stage perception pipeline. Our pipeline leverages a global coordination setup to reduce some practical limitations of frame-based \cite{10341977} and event-based approaches \cite{10161392}.

\subsubsection{Globally Coordinated Multi-Camera Framework}
We deploy an Intel RealSense D415 depth camera near the release point of the tennis ball and a fisheye camera to track the quadruped robot via AprilTags. The ground AprilTag defines the world origin, while the onboard AprilTag provides robot position states $\boldsymbol{p}_r^t$. The transformation from camera to world frame is defined as
\begin{equation}
\boldsymbol{p}_{\text{world}} = T_{\text{cam}}^{\text{world}} \cdot \boldsymbol{p}_{\text{cam}}
\end{equation}
To deal with the jitter of AprilTag between frames and raise the agility of our project, we average the basis vectors and origin position across a sliding window before orthonormalization. This yields a temporally stable transformation matrix. Compared to an onboard binocular depth camera, this design provides stable off-board localization in our workspace, although it constrains deployment to the external-camera field of view.

\subsubsection{Lightweight Object Detection and Throw Event Triggering} 
Dynamic objects such as tennis balls are detected using a fine-tuned YOLOv13n \cite{lei2025yolov13realtimeobjectdetection} model trained on thousands of images under various lighting and motion conditions. To further improve robustness, a simple HSV threshold-based detector runs in parallel as a fallback. This hybrid design ensures reliable detections across cluttered scenes while maintaining inference latency below 10 ms per frame.
Once detected, a finite-state machine (FSM) monitors velocity and displacement signals to identify the actual throwing event. A transition from State 0 \texttt{WAITING} to State 1 \texttt{THROWN} occurs only when
\begin{equation}
   C_{0\to1} = C_{\text{Zvel}} \wedge C_{\text{height}} \wedge C_{\text{data-valid}}
\end{equation}
where $C_{\text{Zvel}}:v_z > v_{\text{thresh}}$, $C_{\text{height}}:z > h_{\text{thresh}}$ and $C_{\text{data-valid}}:\Delta t < \Delta t_{\text{thresh}}$. 
This prevents false positives caused by small jitters, while preserving a simple low-speed triggering rule.

\subsubsection{Trajectory and Landing Position and Timing Prediction}
After the object thrown is confirmed, a Kalman Filter estimates the object’s 3D position $\boldsymbol{p}_b^{t,t-1,...}$ and velocity under a parabolic motion model:
\begin{equation}
    \boldsymbol{x}_{t+1} = \boldsymbol{Ax}_t + \boldsymbol{Bu}_t + \boldsymbol{w}_t, \quad \boldsymbol{z}_t = \boldsymbol{Hx}_t + \boldsymbol{v}_t
\end{equation}
where $\boldsymbol{w}_t$ and $\boldsymbol{v}_t$ is the process and measurement noise, $\boldsymbol{x}_t=[\boldsymbol{p},\boldsymbol{v}]^T$ the state vector with position and velocity, and $\boldsymbol{z}_t=[\boldsymbol{x},\boldsymbol{y},\boldsymbol{z}]^T$ the measurement vector. Even when YOLO temporarily misses detections, the filter maintains stable state estimates. 

\subsubsection{Integration with RL Policy}The perception pipeline, as shown in Fig.~\ref{fig:pipeline}, outputs three key variables: the robot position in the world frame $\boldsymbol{p}_r^t$, the predicted landing coordinates $\boldsymbol{p}_t^{t}$, and the estimated landing time $\text{t}_t$. We would get $\Delta \boldsymbol{p}^t$ from the difference between the robot position and the landing point coordinates. $\Delta \boldsymbol{p}^t$ and $\text{t}_t$ are transmitted via LCM to the Jetson Orin Nano onboard and form part of the observation space for the RL policy. By conditioning the policy on both spatial and temporal predictions, the quadruped can align its arrival time with the object’s landing while preserving stable locomotion.

%% file: source/rl_part.tex
\subsection{Reinforcement Learning for Position-conditioned Locomotion}
 
\begin{figure*}[!htbp]
\centering
    \includegraphics[width=0.8\textwidth]{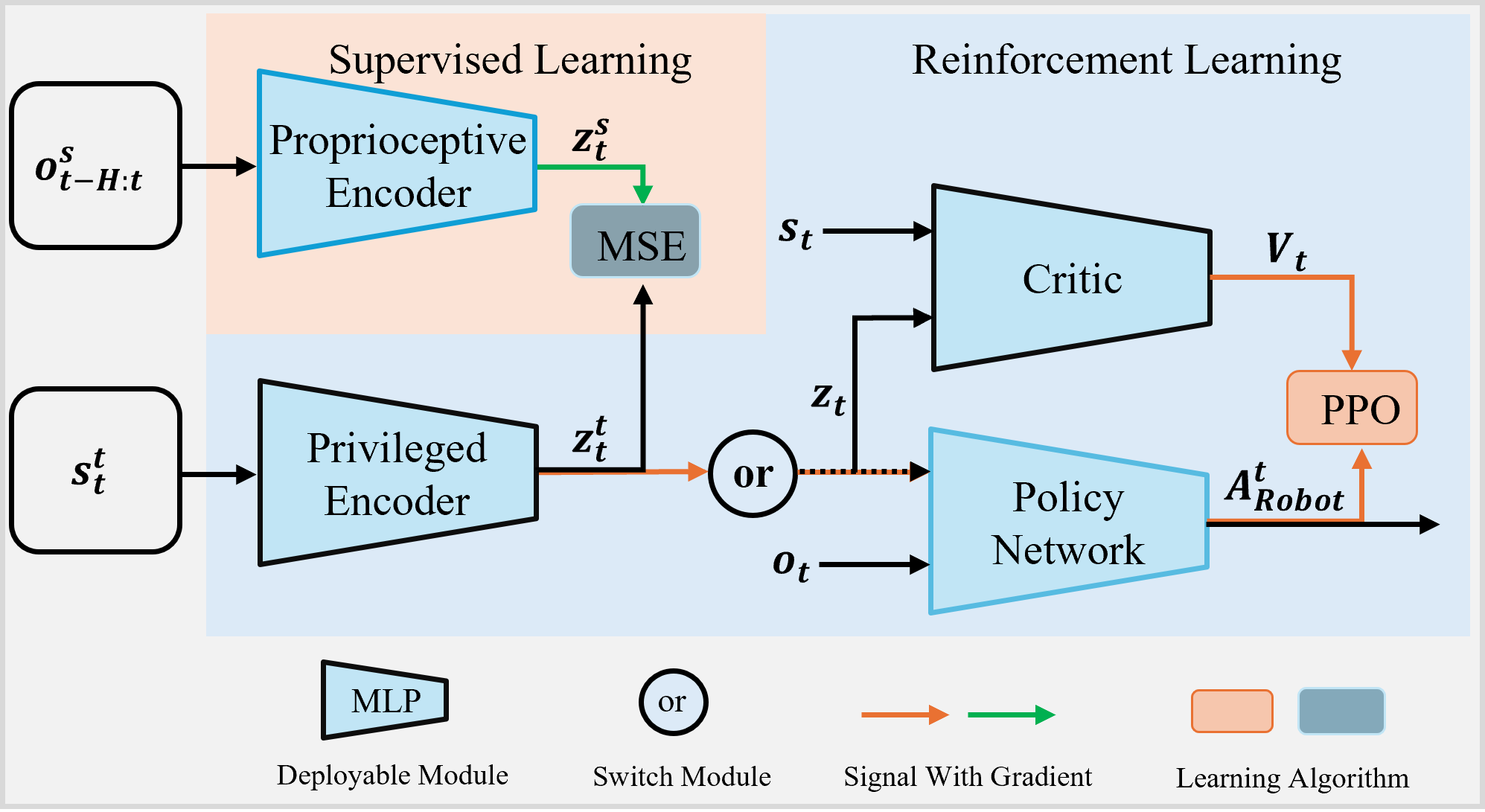}
    \caption{Overview of the proposed position-conditioned local navigation policy training framework. We first train a teacher policy with full observability under PPO, and a student policy is trained in a supervised manner with only the proprioceptive observation to enable the sim-to-real transfer}
    \label{fig:Teacher-Student}
\end{figure*}

While direct Proximal Policy Optimization (PPO) \cite{schulman2017proximalpolicyoptimizationalgorithms} training is often sufficient for basic, steady-state locomotion, our time-constrained interception task demands rapid target reaching. High-acceleration maneuvers, such as rapid turning and sprinting, are highly sensitive to unobservable environmental dynamics like ground friction, precise contact forces, and varying motor strengths. To improve robustness under these factors, we adopt a concurrent teacher-student distillation framework following \cite{10670293}, as illustrated in Fig.~\ref{fig:Teacher-Student}. The teacher policy uses privileged simulation information during training, while the deployed student policy relies only on deployable proprioceptive observations and task commands. This architecture is used as a practical sim-to-real training choice rather than as a claim that direct PPO is fundamentally unsuitable for this task. A separate comparison with direct PPO under the same position-and-time command interface would be required to isolate the effect of distillation, which we leave for future work. We construct the parallel simulation environment for RL training with the IsaacGym simulator \cite{makoviychuk2isaac}.
\subsubsection{Observation and Action Space}
We denote the \( \bm{o}_t \in \mathbb{R}^{46} \) in Fig.~\ref{fig:Teacher-Student}, as the observation of the agent consisting of the angular velocity components of the robot base in the world coordinate system, \( \bm{\omega}_t \in \mathbb{R}^3 \); the projection of the gravitational vector in the robot coordinate system, \( \mathbf{g}_{proj} \in \mathbb{R}^3 \); objective commands, \( \mathbf{c} \in \mathbb{R}^3 \); normalized remaining time, \( t_{t} \in \mathbb{R}^1 \); the deviation between the current joint position of each degree of freedom and the default position, \( \Delta \mathbf{q} \in \mathbb{R}^{12} \); joint velocities of each degree of freedom, \( \mathbf{\dot{q}} \in \mathbb{R}^{12} \); and the action of the previous time \( \mathbf{a}_{t-1} \in \mathbb{R}^{12} \). The commands include the target position error $(\Delta x, \Delta y)$ and orientation error $(\Delta \theta)$, which are calculated by the position difference between the robot position and predicted ball landing point, $\boldsymbol{p}_t^t$, as shown in Fig.~\ref{fig:pipeline}. This command can guide the robot to move to the target. In addition, the previous action is the proprioception and $\bm{o}_{t-H:t}$: as its history, where $H$ represents the history length. To avoid signal delay caused by an overly large input historical dimension, we set $H=5$. The privileged observation, \( \bm{s}_t \in \mathbb{R}^{148} \), consists of $\bm{o}_t$, joint torques and accelerations, body linear velocity, feet contact forces, PD control gains, friction coefficients, motor strength, leg mass, and base mass. The policy network also receives the compressed latent representation of the full robot states, \( \bm{z}_t^s \in \mathbb{R}^{32} \), which is the reconstructed latent vector of the normalized latent representation $\bm{z}_t$. $\bm{V}_t$ refers to the value estimation by the critic network. With the observation input, the policy will output the action, $\bm{a}_t$ in Fig.~\ref{fig:pipeline}, to the robot to complete the time-constrained position-reaching task.

\begin{table}
\TBL{\caption{Reward Function Elements\label{table_reward}}}
{\begin{fntable}\centering
\begin{tabular}{llc} 

\hline
Reward & Equation($r_i$) & Weight($w_i$) \\
\hline
Lin. velocity tracking & $\exp\left(-10\Vert \bm{c}_{xy} - \bm{v}_{xy}\Vert^2\right)$ & $1.0$\\ 
Ang. velocity tracking & $\exp\left(-10( c_{z} - \omega_z )^2\right)$ & $0.5$\\ 
Pos. tracking & $\exp(-10\bm{c}_{xy}^2)$ & 1.0\\
Yaw tracking & $\exp(-10c_z^2)$ & 0.5\\
Linear velocity (z) & $v_{z}^2$ & -2.0\\
Angular velocity (xy) & $\bm{\omega}_{xy}^2$ & -0.05\\
Orientation & $(g_x^{\text{proj}})^2 + (g_y^{\text{proj}})^2$ & -40.0\\
Joint accelerations & $(\frac{\mathbf{\dot q}_{t}-\mathbf{\dot q}_{t-1}}{\Delta t})^{2}$ & $-4 \times 10^{-7}$\\
Joint power & $\big|\mathbf{\tau}\big| \times \big|\mathbf{\dot q}\big|$ & $-2 \times 10^{-4}$\\
Collision & $\sum_k \mathbf{1}\{\|\mathbf{f}_k\|>0.1\}$ & -1.0\\
Action rate & $(\mathbf{a}_{t-1}-\mathbf{a}_{t})^{2}$ & -0.01\\
Action smoothness & $(\mathbf{a}_{t} - 2\mathbf{a}_{t-1} - \mathbf{a}_{t-2})^{2}$ & $-2 \times 10^{-3}$\\
Torque & $\bm{\tau}_i^{2}$ & $-2 \times 10^{-4}$\\
Joint position limits & $\Delta\mathbf{q}$ & -10.0\\
Joint velocity & $\mathbf{\dot{q}}^{2}$ & $-1.5 \times 10^{-3}$\\
Stand still position & $|\mathbf{q}_i - \mathbf{q}_{i,\text{default}}|$, if dist. $<0.1$ & -10.0\\
Base height & $(z_{\text{root}} - z_{\text{target}})^{2}$ & -10.0\\
Feet air time & $r_{ft}$ & 3.0\\
Reach Target & $r_{pos\_time}$ & 30\\
Task & $r_{t}$ & 100\\
Feet acceleration & $\Delta{v_{f}}^2$ & $-1 \times 10^{-4}$\\
Exploration & $r_{e}$ & 1\\
Stalling penalty & $r_{sp}$ & 1\\
Stop yaw velocity & $r_{sy}$ & -0.1\\
feet height & $\exp(-10(h_f-0.1)^2)$ & -2.0 \\
\hline
\end{tabular}
\end{fntable}}
\end{table}

\subsubsection{Time-Aware Reward Design}
Table \ref{table_reward} shows the reward terms we used. Most task reward functions are about spatial tracking and time alignment, where the detailed formulation follows \cite{doi:10.1126/scirobotics.adi7566}. In this way, we can explicitly address the temporal misalignment issue highlighted in prior works. We use a time-gated reward structure based on a duration mask, defined as $M(t_{go}, D) = \frac{1}{D} \mathbb{I}(t_{go} \le D)$, where $t_{go}$ is the remaining time to impact and $D$ is a specific duration threshold \cite{he2024agilesafelearningcollisionfree}. This effectively scales the spatial tracking rewards based on urgency. The time-to-impact reward is formulated as:
\begin{equation}
r_{pos\_time} = M(t_{go}, D_{rew}) \cdot \frac{1}{1 + \left( \frac{\|p_{target} - p_{robot}\|}{\sigma_p} \right)^2}
\end{equation}
Unlike standard spatial tracking, this formulation encourages the agent to manage its arrival time, heavily rewarding the robot only when it is spatially accurate and temporally aligned with the ball's predicted landing window. Since the reward combines several task, stability, and exploration terms, we interpret the observed motion pattern as the outcome of the overall training design rather than attributing it to a single reward component.

Safety and smoothness terms refer \cite{11128639}, \cite{doi:10.1126/scirobotics.adi7566}. We normalize each term to comparable ranges and weight task terms higher than auxiliaries, but keep auxiliary weights sufficiently large to prevent catastrophic behaviors, such as falling, sinking, and excessive torques. Notably, there is no specific velocity direction term \cite{he2024agilesafelearningcollisionfree} to prescribe a two-phase rotate-and-run sequence. However, as shown in the experiments section, a rotation-prioritized behavior is observed in some trials, where the trained policy drives the robot rapidly, turning to the direction where the target position is at first and runs towards the position, as displayed in Fig. ~\ref{fig:exp_emergent}. This observed behavior suggests responsive time-aware position-conditioned locomotion training fashion. In addition, to alleviate the sim-to-real gap and enhance the robustness of our policy, we implement the domain randomization for the training, where the detailed terms are shown in Table \ref{domain_rand}.
\begin{table}
\TBL{\caption{Domain Randomization Terms\label{domain_rand}}}
{\begin{fntable}\centering
\begin{tabular}{lcc} 
\hline
Randomization Term & Range & Unit \\
\hline
Friction & $[0.5, 1.25]$  & - \\
Base mass offset $\Delta m$ & $[-1.0, 1.0]$ & kg \\
Robot push & $[-1.0, 1.0]$ $/15s$ & m/s \\
CoM of base  & $[-1.0, 1.0]^3$  & cm   \\
Motor strength factor & $[0.9, 1.1]$   & - \\
Motor offset & $[-2.0, 2.0]$   & cm \\
Joint $K_p$ factor  & $[0.8, 1.2]$   & N$\cdot$rad   \\
Joint $K_d$ factor  & $[0.8, 1.2]$    & N$\cdot$rad/s \\
\hline
\end{tabular}
\end{fntable}}
\end{table}

\subsubsection{Curriculum Learning Framework}
To enhance the Learning efficiency of the agent, we adopted the \textit{Curriculum Learning} strategy and carried out progressive expansion for the range of motion instructions, such as target position and orientation. In the initial stage, the distance between body position and target point, and the orientation of the target point are uniformly and randomly sampled from the range $[-0.5, 1.0]$ m and $[-1.5, 1.5]$ rad, respectively. If trained within this initial range, the robot will not exhibit abnormal movements, such as turning around and walking backwards. As the performance of the agent improves, the instruction range gradually expands to  $[-5.0, 7.5]$ m and $[-\pi, \pi]$ rad. Specifically, when the agent performs well within the current instruction range, the sampling interval of the target distance and angle is automatically increased to promote its exploration of a wider range of motion capabilities. Moreover, to encourage exploration of task-relevant motion strategies, we employ the Random Network Distillation technique (RND) as the intrinsic reward \cite{burdaexploration} to encourage exploration for the state-action space. RND is used as an exploration aid in the training pipeline; isolating its individual effect is beyond the scope of this system-level evaluation.

%% file: source/experiment.tex
\section{Experiment}
In this section, we compare our approach to a velocity tracking baseline in both simulation and physical sim-to-real experiments, and to an instantaneous ball-state (IBS) baseline solely in simulation. 
Both baselines are trained under the identical domain randomization settings detailed in Table \ref{domain_rand} and utilize the reward terms from Table \ref{table_reward}. To ensure a controlled simulation comparison, the reward scales for each baseline were meticulously tuned via an extensive hyperparameter search to achieve their best possible performance. Specifically, for the IBS baseline, we train the policy by sending the ball's current frame position and velocity directly into the robot's observation space. We call this the IBS baseline to distinguish it from impact-point-conditioned policies that receive a predicted landing target as input; the IBS baseline instead receives current ball position and velocity. For the other baseline, we use the linear controller defined as the following equations, to guide the robot to walk towards the target point,
\begin{equation}
K_p(\bm{P}_d-\bm{P}_r) = \bm{v}_{xy}^{cmd}
\end{equation}
\begin{equation}
K_p(\theta_d-\theta_r) = \omega^{cmd}
\end{equation}
where $K_p$ is a coefficient; $\bm{P}_d$, $\theta_d$ are desired position and angle; $\bm{P}_r$, $\theta_r$ are the current robot position and orientation. After tuning, we find that $K_p=2.5$ has the best performance for the velocity tracking approach. Our position-conditioned method is achieved by providing task-related rewards in RL training without any further tuning.

\subsection{System Setup}

The hardware setup consists of a Unitree Go2 quadruped robot with a $22 \times 28 \times 9$ cm$^3$ size basket, which means that the acceptable error range for us is within 17.8 cm, and an onboard NVIDIA Jetson Orin Nano for locomotion policy inference. The perception system utilizes two stationary cameras, explicitly forming a decoupled, off-board perception setup. 
\begin{figure}[!htbp]
    \centering
    \includegraphics[width=0.8\linewidth]{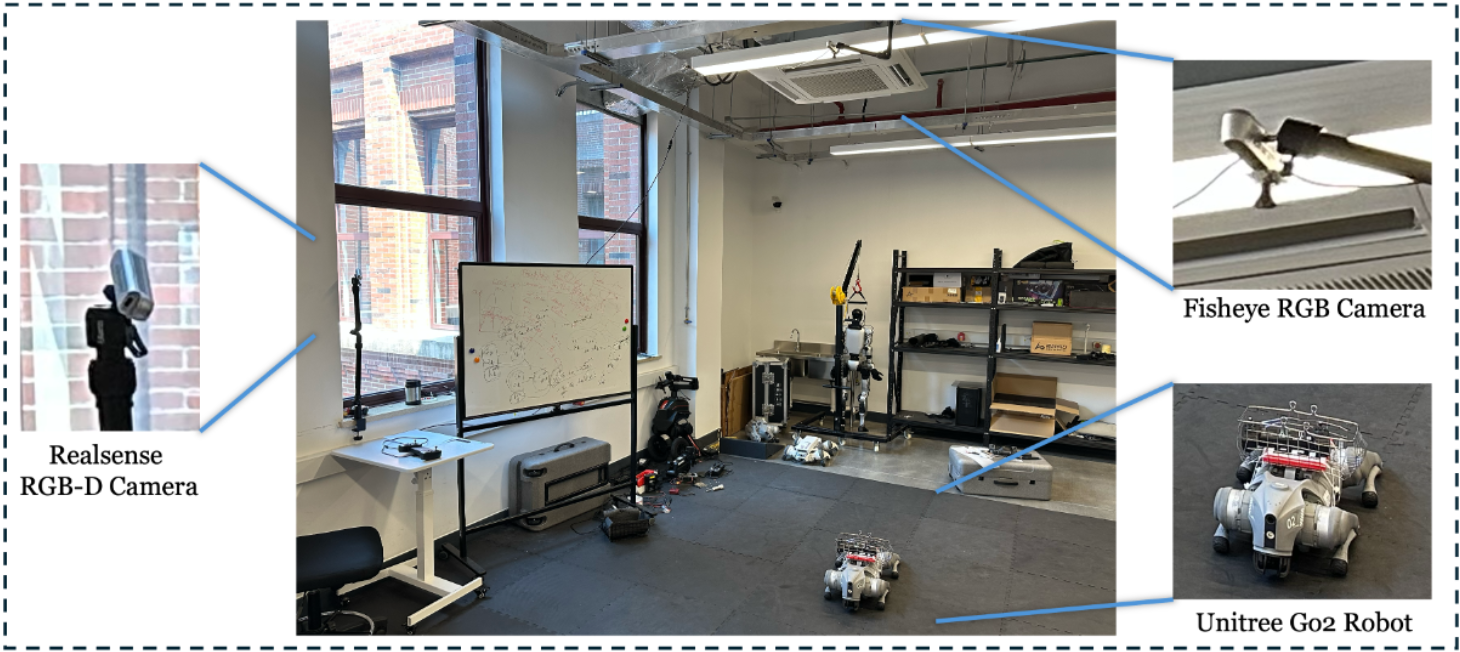}
    \caption{Experiment Setup}
    \label{fig:exp_setup}
\end{figure}

\begin{figure*}[!htbp]
\centering
    \includegraphics[width=\textwidth]{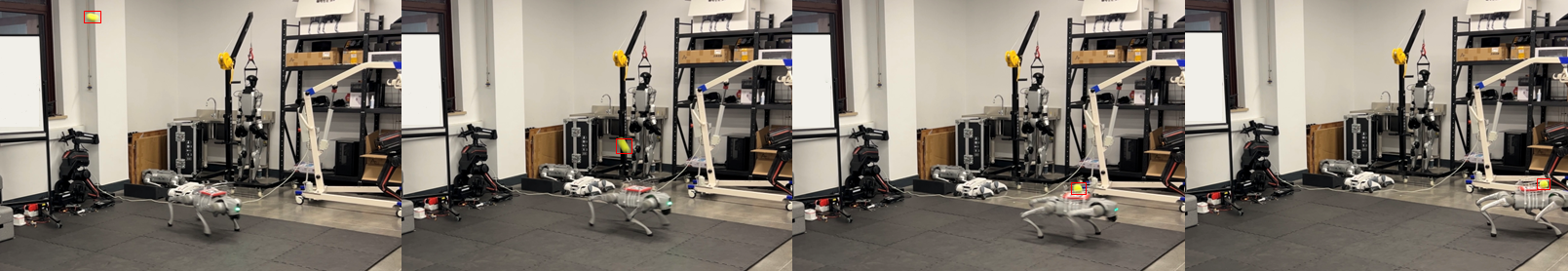}
    \caption{Sim-to-real tennis ball catching experiment}
    \label{fig:Sim2Real}
\end{figure*}
An overhead RGB fisheye camera, calibrated for undistortion, provides a global view for localizing the robot via an AprilTag on its back. A second Intel RealSense D415 camera is positioned near the ball's release point, with its field of view encompassing the flight path and the base coordinate system defined by the AprilTag. The whole experiment setup is illustrated in Fig. ~\ref {fig:exp_setup}.


The software pipeline on an external PC uses YOLOv13n to detect the ball from the RealSense D415 feed. A basic Kalman filter with a state vector of $(x, y, z, \dot{x}, \dot{y}, \dot{z})$ and positional measurements $(x, y, z)$ as inputs is used for trajectory prediction, estimating the ball's landing point and time.

The robot's onboard computer receives the predicted target landing point and time-to-impact. The control policy subsequently uses this target point, alongside the robot's proprioceptive data, to generate the locomotion commands for interception.

\subsection{Evaluation with Simulation}
We conducted the simulation evaluation in MuJoCo to preliminarily test our method and the baseline. To simulate the ball-throwing task, we use the oblique projectile model and constrain the landing point within the $2$ m range of the robot with a time limit between $1.15$ and $1.35$ seconds. This simulated flight-time range is different from the real-world trials, so the two evaluations are complementary rather than strictly like-for-like. The landing point of the ball in the simulation is shown in Fig. \ref{fig:landing_points}, which is all around the robot.

\begin{table}
\TBL{\caption{Simulation Catch Success Rate (Catch S.R.)\label{tab:sim_sr}}}
{\begin{fntable}\centering
\begin{tabular}{c c c c}
\hline
Catch \textbf{S.R.} (\%) & $<0.5$ & 0.5--1 & 1--2\\
\hline
velocity tracking & 67.39\% & 67.11\% & 16.28\% \\
 IBS baseline & 18.33\% & 11.32\% & 2.24\% \\
time-conditioned target-reaching & 86.73\% & 22.45\% & 35.63\% \\
\hline
\end{tabular}
\end{fntable}}
\end{table}

\begin{figure}[!htbp]
    \centering
    \includegraphics[width=0.7\linewidth]{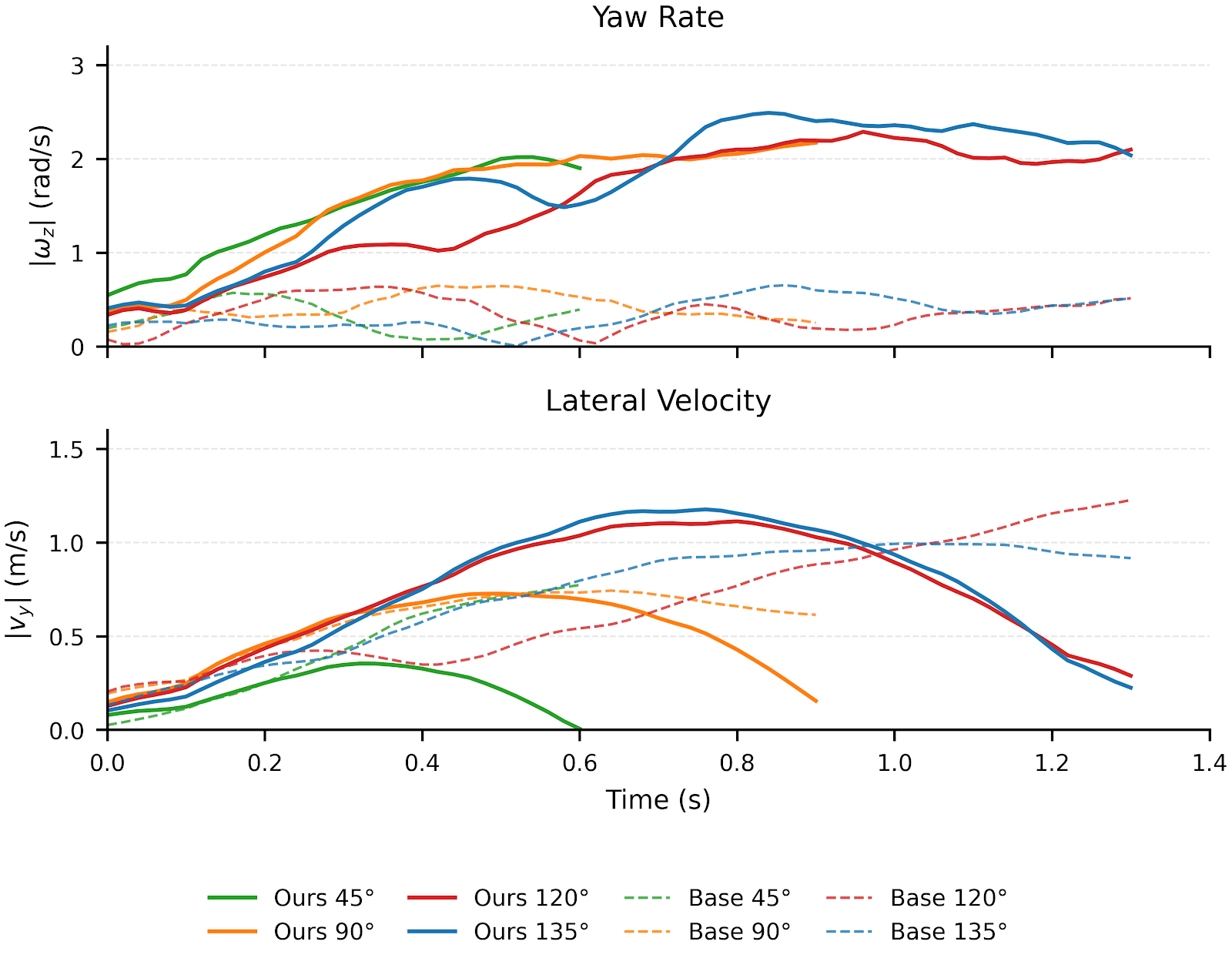}
    \caption{Comparison of angular and lateral velocities. Ours executes a rapid turn with minimal lateral slip, whereas the baseline exhibits persistent lateral sliding and low rotation}
    \label{fig:Yaw_velocity}
\end{figure}

We conducted 300 simulation experiments for each method. We define successful catching as the ball being thrown into the basket on the back of the robot, including those that are thrown into the basket and then bounce out. To obtain the velocity profiles shown in Fig. \ref{fig:Yaw_velocity}, we fixed the robot's initial position and orientation, and threw the ball to land at a distance of $2$ m at various angles relative to the robot.
The result of the simulation experiments is shown in Table \ref{tab:sim_sr}. The IBS baseline's poor performance highlights the inefficiency of implicitly learning ballistic dynamics and precise timing directly from instantaneous frame-wise observations, but it should not be interpreted as a comparison with a goal-conditioned impact-point baseline. Also, we find that the extreme ball catching distance of our method is $1.78$ m and $1.37$ m for the baseline. As illustrated in Fig. \ref{fig:Yaw_velocity}, our policy exhibits a rotation-prioritized action: the robot first performs a rapid turn with high angular velocity and then rushes straight toward the target. In contrast, the baseline method primarily relies on lateral movement with significantly lower angular velocity. These tables show that our method performs better ball-catching ability in short range and long range, and omnidirectional mobility capability. Our method also has higher success rates in the tested short- and long-distance bins compared with the baseline. However, the baseline performs better in the medium range. Because our policy prefers the robot to turn at first, then rush towards the target position. But the velocity tracking approach leads the robot to move laterally to the target position when the target position is on the side of the robot. In the short range, the command of baseline is very small, which makes the robot move slowly or even stand still. In the medium range, the laterally moving time is shorter than the sum of the turning and rushing forward time. In the larger run, turning and rushing is more effective for the robot to complete the tight time-constrained task.

\begin{figure}[!hbtp]
    \centering
    \includegraphics[width=0.7\linewidth]{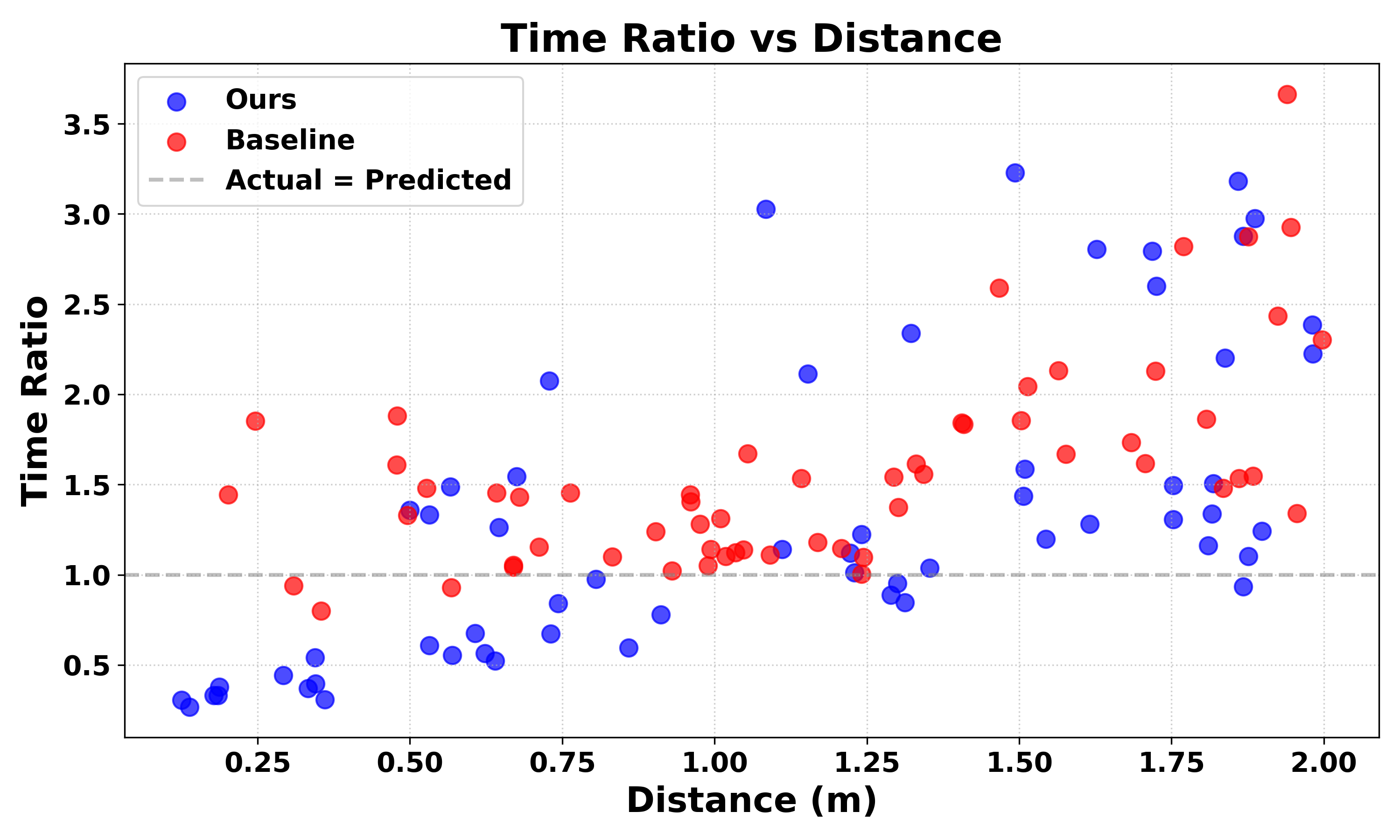}
    \caption{Comparison of Time Ratio ($t_{actual}/t_{predicted}$) across catching distances. Time ratios $< 1.0$, $= 1.0$, and $> 1.0$ signify preemptive agile arrival, perfect synchronization, and late arrival, respectively}
    \label{fig:time_comparison}
\end{figure}

To further quantify the timing behavior and directly address the timing synchronization of the policy, we evaluate the Time Ratio across varying catching distances, as depicted in Fig. \ref{fig:time_comparison}. The results suggest the usefulness of our time-aware reward formulation. Our position-conditioned method consistently achieves a lower Time Ratio than the baseline. Specifically, at shorter distances, under $1.0$ m, our policy predominantly yields ratios below $1.0$, enabling the robot to reach the target early and stabilize before impact. Conversely, the baseline frequently yields ratios above $1.0$, struggling to synchronize with the temporal constraints. Even at extreme distances, $1.0$ m to $2.0$ m, where the task demands longer-range rapid reaching, our approach maintains a significantly lower and tighter Time Ratio distribution. This quantitative evidence suggests that explicit time conditioning can help coordinate arrival timing in these trials.

\subsection{Real World Validation}

\begin{figure}[!htbp]
    \centering
    \includegraphics[width=0.8\linewidth]{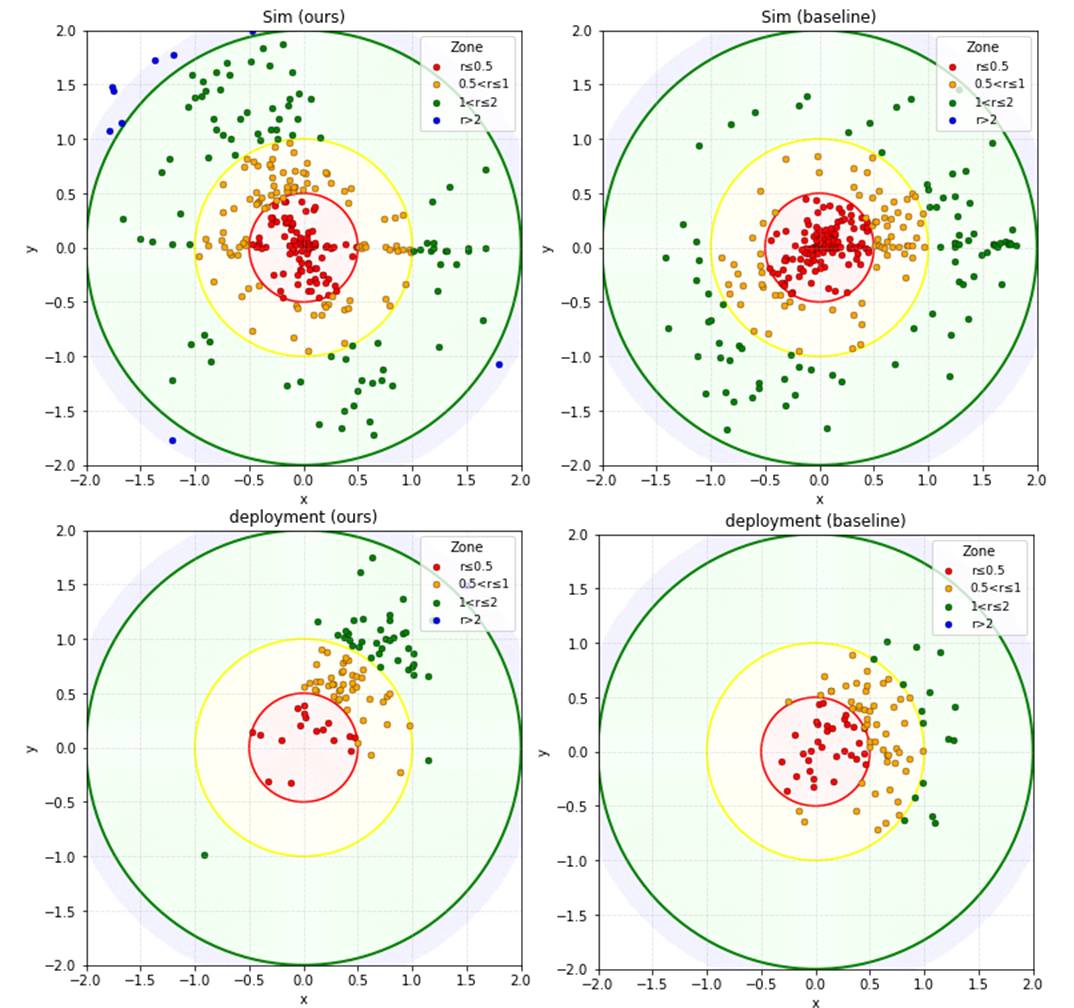}
    \caption{Distribution of landing points. Comparison of landing positions for our method (left) and the baseline (right) in both simulation (top row) and real-world (bottom row). Note that the robot is facing along the $y$ direction}
    \label{fig:landing_points}
\end{figure}

To assess the hardware deployment of our approach, we deployed both our policy and the baseline policy on a Unitree Go2 robot. In the simulation test, the velocity-tracking baseline achieved its best performance with $k_p$=2.5. However, we found this setting caused insufficient actuation for the robot to move when catching a ball at a short distance, as mentioned in the simulation experiment. Therefore, during deployment, we adjusted $k_p$=5.0 when the distance between the target position and robot position was smaller than 0.5 m to improve the baseline’s short-distance responsiveness without changing its policy structure. In addition, we apply the same yaw-command shaping rule to both deployed methods: when the desired turning angle is larger than $60^\circ$, we give a $120^\circ$ yaw command to make turning faster. Since this rule is shared by both methods, it is not a method-specific advantage of our policy. Before the formal experiments, we conducted preliminary experiments. We found that for our policy, the robot performed better with landing points to its front-lateral side, whereas lateral landing points were easier to catch for the velocity-tracking method.
\begin{figure}[!htbp]
    \centering
    \includegraphics[width=0.8\linewidth]{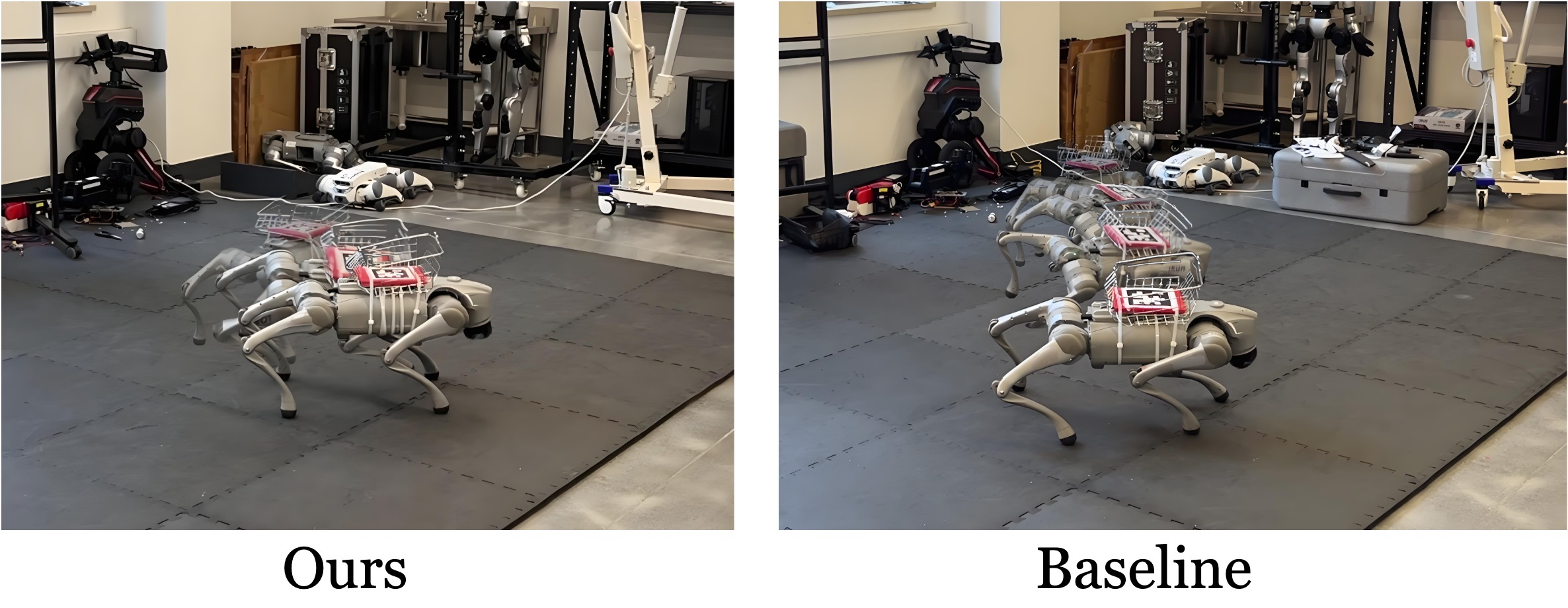}
    \caption{Comparison of different motion patterns. Our method demonstrates a rotation-prioritized locomotion pattern, while the baseline makes the robot move laterally}
    \label{fig:exp_emergent}
\end{figure}
We fixed the original robot orientation forward, and threw a tennis ball randomly within a 0–2 m range near their \textit{feasible catching zones}, between 0.8 and 1.2 s. Thus, the real-world comparison is feasibility-oriented rather than a fully standardized workspace-wide benchmark. Each policy was tested for 100 trials with the same robot and perception pipeline, using the same models that achieved the best performance in simulation. Experiments were grouped by landing distance ($<$0.5 m, 0.5–1.0 m, and 1.0–2.0 m). The distribution of ball landing points across all trials is shown in Fig. \ref{fig:landing_points}, confirming that throws were randomized within the designated 0–2 m range.

\begin{table}
\TBL{\caption{Sim-to-real Success Rate (S.R.) Comparison\label{tab:real_sr}}}
{\begin{fntable}\centering
\begin{tabular}{c c c c}
\hline
Track \textbf{S.R.} (\%) & $<0.5$ m & 0.5--1 m & 1--2 m\\
\hline
baseline & 25.00\% & 51.92\% & 43.75\% \\
\textbf{ours} & \textbf{62.50}\% & \textbf{64.44}\% & \textbf{63.16}\% \\
\hline
\hline
Catch \textbf{S.R.} (\%) & $<0.5$ m & 0.5--1 m & 1--2 m\\
\hline
baseline & 6.25\% & 9.62\% & 25.00\% \\
\textbf{ours} & \textbf{37.50}\% & \textbf{24.44}\% & \textbf{36.84}\% \\
\hline
\end{tabular}
\end{fntable}}
\end{table}

The results are summarized in Table \ref{tab:real_sr}, organized separately by strict criterion and tracking criterion. When rim contacts are considered successful, which is reflected as the track S.R., the position-conditioned policy S.R. consistently exceeds 60\%, while the baseline performance is lower. Track S.R. is a diagnostic contact metric rather than a strict catch metric. Under the strict definition, where only direct hitting the basket bottom is counted as successful, that is the catch S.R., the performance gap becomes even more significant: the velocity-tracking baseline achieves less than 10\% success rates in $<$0.5 m and 0.5 m - 1 m cases, and gets 25\% success rates in 1 m - 2 m case, essentially failing to catch in real-world settings. In addition, the extreme ball catching distance of our method is 1.70 m, and 1.23m for the baseline. The sim-to-real gap may be due to the soft terrain, which makes the robot encounter considerable resistance when turning around quickly. Also, there exist the inevitable problems, such as communication delay, motor strength variation due to power voltage changing, and so on. A detailed decomposition of detection, prediction, latency, and tracking errors is left for future work.
These findings suggest that while the velocity-conditioned baseline degrades dramatically during sim-to-real transfer, our position-conditioned approach maintains its advantage in real deployments despite uncertainties. This highlights that explicit position information is useful for robust physical interception.
In conclusion, extensive simulations and real-world experiments validate that our integrated perception-informed locomotion framework achieves higher catching rates under the tested setup compared to the baseline.

%% file: source/conclusion.tex
\section{Conclusion and Future Work}
We proposed a position-target driven framework enabling dynamic quadruped ball catching with higher success rates than the velocity-tracking approach under our tested setup.
A vision module detects the ball, while a Kalman Filter predicts its landing position and timing to provide spatio-temporal inputs for locomotion. By decoupling perception and control, we trained a position-conditioned local navigation policy via RL for hardware deployment.
Our decoupled design enables dynamic interception by providing predicted landing-position and landing-time commands to the policy. By employing a Kalman Filter to inject Newtonian priors, the system helps maintain trajectory estimates during temporary visual occlusions, reducing reliance on learning ballistic dynamics directly from raw kinematic observations.
Real-world deployments and sim-to-sim evaluation show that our framework achieves higher catching rates than the velocity-tracking baseline during sim-to-real transfer.
Integrating such standard modules can produce a rotation-prioritized locomotion pattern. The results indicate that
our approach achieved higher rates than the velocity-tracking baseline and was deployable on hardware under our experimental setup. 
While the policy exhibits dynamic locomotion, the absolute real-world catching success rate remains modest. Compounding factors such as hardware actuation latency and visual tracking noise currently limit overall system robustness. Furthermore, our current agile local navigation policy is primarily evaluated on flat terrain, and the robot is constrained within the field of view of a global camera. The current experiments also do not isolate the effects of time conditioning, teacher-student distillation, reward terms, or prediction error. Future work will focus on refining the sim-to-real transfer to improve the success rate, extending the agile interception framework to complex terrains, and enabling onboard perception to remove the external camera restriction, thereby improving the system's adaptability and scalability in unstructured environments.